\documentclass[letterpaper, 10 pt, conference]{ieeeconf}
\IEEEoverridecommandlockouts 

\usepackage{amsmath}
\usepackage{amssymb}
\usepackage{xcolor}
\usepackage{cite}
\usepackage[colorlinks=true, citecolor=cyan]{hyperref}
\usepackage{graphicx}
\usepackage{float}
\usepackage{multirow}
\usepackage{soul}
\usepackage{booktabs}
\usepackage[table]{xcolor}
\usepackage{mdframed}

\DeclareMathOperator*{\argmin}{arg\,min}

\title{\LARGE \bf
From Obstacles to Etiquette: Robot Social Navigation with VLM-Informed Path Selection
}

\author{Zilin Fang$^1$, Anxing Xiao$^1$, David Hsu$^{1,2,*}$, and Gim Hee Lee$^{1,*}$
\thanks{This research is supported by Agency for Science, Technology \& Research (A*STAR), Singapore, under its National Robotics Program (No. M23NBK0053), and the National Research Foundation (NRF) Singapore, under its NRF-Investigatorship Programme (Award ID. NRFNRFI09-0008).
}
\thanks{$^{1}$School of Computing, $^{2}$Smart Systems Institute, National University of Singapore, Singapore.
        {\tt\footnotesize \{zilin.fang, anxingxiao\}@u.nus.edu, dyhsu@comp.nus.edu.sg, gimhee.lee@nus.edu.sg}}%
\thanks{$^{*}$ Co-supervision}
}

\begin{document}

\maketitle
\thispagestyle{empty}
\pagestyle{empty}

\begin{abstract}

Navigating socially in human environments requires more than satisfying geometric constraints, as collision-free paths may still interfere with ongoing activities or conflict with social norms. Addressing this challenge calls for analyzing interactions between agents and incorporating common-sense reasoning into planning.
This paper presents a social robot navigation framework that integrates geometric planning with contextual social reasoning. The system first extracts obstacles and human dynamics to generate geometrically feasible candidate paths, then leverages a fine-tuned vision-language model (VLM) to evaluate these paths, informed by contextually grounded social expectations, selecting a socially optimized path for the controller. This task-specific VLM distills social reasoning from large foundation models into a smaller and efficient model, allowing the framework to perform real-time adaptation in diverse human–robot interaction contexts.
Experiments in four social navigation contexts demonstrate that our method achieves the best overall performance with the lowest personal space violation duration, the minimal pedestrian-facing time, and no social zone intrusions. Project page: \href{https://path-etiquette.github.io}{\textcolor{cyan}{path-etiquette.github.io}}
\end{abstract}

\section{Introduction}

\begin{figure}[tb]
  \centering
  \includegraphics[width=0.9\linewidth]{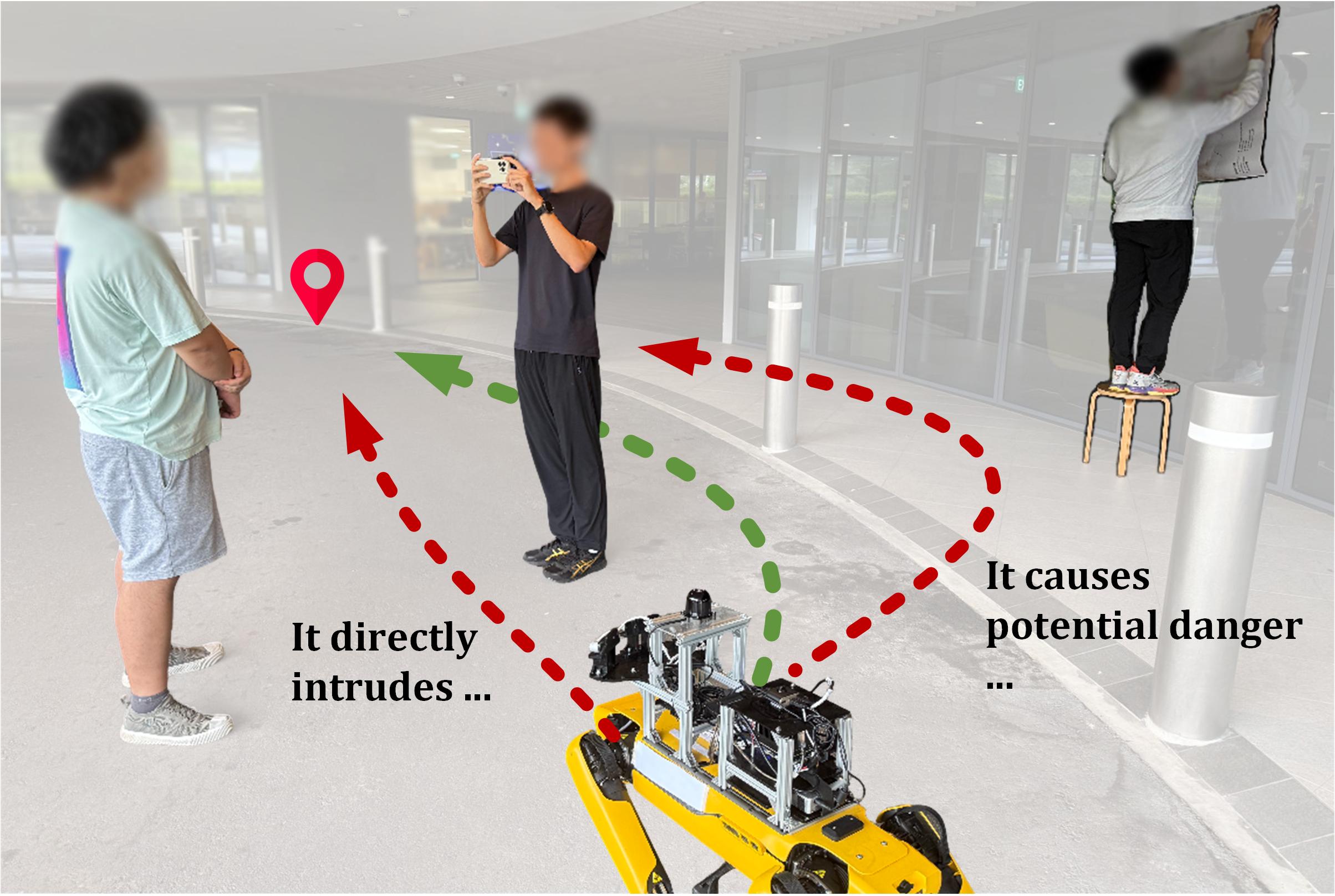}
  \vspace{-0.8em}
  \caption{Illustration of robot navigation in a scenario with three geometrically feasible sampled paths, where the robot should reason about social conventions to select the most appropriate path.}
  \vspace{-1.5em}
  \label{fig:teaser}
\end{figure}

Navigating robots in crowded environments is both challenging and critical for applications such as autonomous delivery and guidance~\cite{triebel2016spencer, xiao2021robotic, cai2024navigating}. 
Beyond goal reaching under geometric constraints, robots should also adhere to social etiquette in crowds—recognizing cues such as people posing for photographs or a worker on a ladder—and act to minimize disruption and risk~\cite{francis2025principles}, as illustrated in Fig.~\ref{fig:teaser}. However, generating such contextually adaptive motions remains difficult, as it requires both spatial and semantic awareness to identify socially compliant spaces, yet relevant robot datasets are scarce. In this work, \textit{we propose a social robot navigation framework that leverages Vision-Language Models (VLMs) to evaluate the social conventions associated with feasible geometric paths}, enabling robots to follow social etiquette without explicitly modeling socially compliant spaces.

Prior works employing VLMs for social robot navigation have largely focused on high-level semantic reasoning at the symbolic level, such as predicting explicit social relationships~\cite{luo2025gson}, choosing navigation goals projected in images~\cite{nasiriany2024pivot, sathyamoorthy2024convoi}, predicting pedestrian intention~\cite{munir2025pedestrian}, or scoring predefined actions~\cite{song2024vlm}. However, evaluating purely on symbolic representations fails to capture real-world execution effects, which can lead to intrusions or navigation failures. In addition, frequent queries to large VLMs can result in slow inference, limiting real-time applicability.

To address the above challenges, we present a navigation framework that integrates spatial constraints from obstacles and humans with semantic adaptations derived from human-environment, human-human, and human-robot interactions in real-time. 
The social robot navigation task is formulated as a multi-objective optimization problem over geometric feasibility and in-context semantics. These objectives are assumed to be decomposable, with the optimal semantic solution space lying within a well-covered subset of the geometric optimum. Accordingly, we adopt a hierarchical architecture with asynchronous modules for geometry-aware path planning, socially compliant path selection, and safe reactive control. 
At the high level, candidate paths are first sampled under geometric constraints, and then VLMs are employed to select grounded, socially compliant paths from this feasible set. This proceeds in a receding-horizon fashion. For real-time inference, we introduce a pipeline that distills social reasoning capabilities from large models into a smaller model. At the low level, the selected path is fed back to the path planning module as a reference for generating new paths, while a modified ORCA algorithm is employed to ensure pedestrian avoidance. At a conceptual level, our propose–select strategy is close to prior work~\cite{song2025vl, liang2025mosu}, but it targets a different problem domain with dynamic, interaction-aware reasoning objectives and a different system design.

We evaluate our method through controlled experiments on a Boston Dynamics Spot legged robot in diverse social scenarios involving human activity. These scenarios are motivated by the insight that, while social behavior is inherently multimodal and shaped by individual preferences, many situations follow common patterns that minimize intrusion into social zones. This design reduces reliance on subjective post-hoc surveys and enables social performance to be quantified using interruption-related metrics. Real-world experiments demonstrate that our approach achieves superior social compliance and more efficient goal-reaching compared to representative baselines, including group-based~\cite{wang2022group, luo2025gson}, RL-based~\cite{liu2023intention}, VLM-based~\cite{song2024vlm} methods, as well as a foundation model for visual navigation~\cite{shah2023vint}.
Ablation studies further show that our approach generates collision-free and socially compliant paths, which direct VLM path prediction cannot reliably guarantee.

\begin{figure*}[t]
  \centering
  \includegraphics[width=0.95\linewidth]{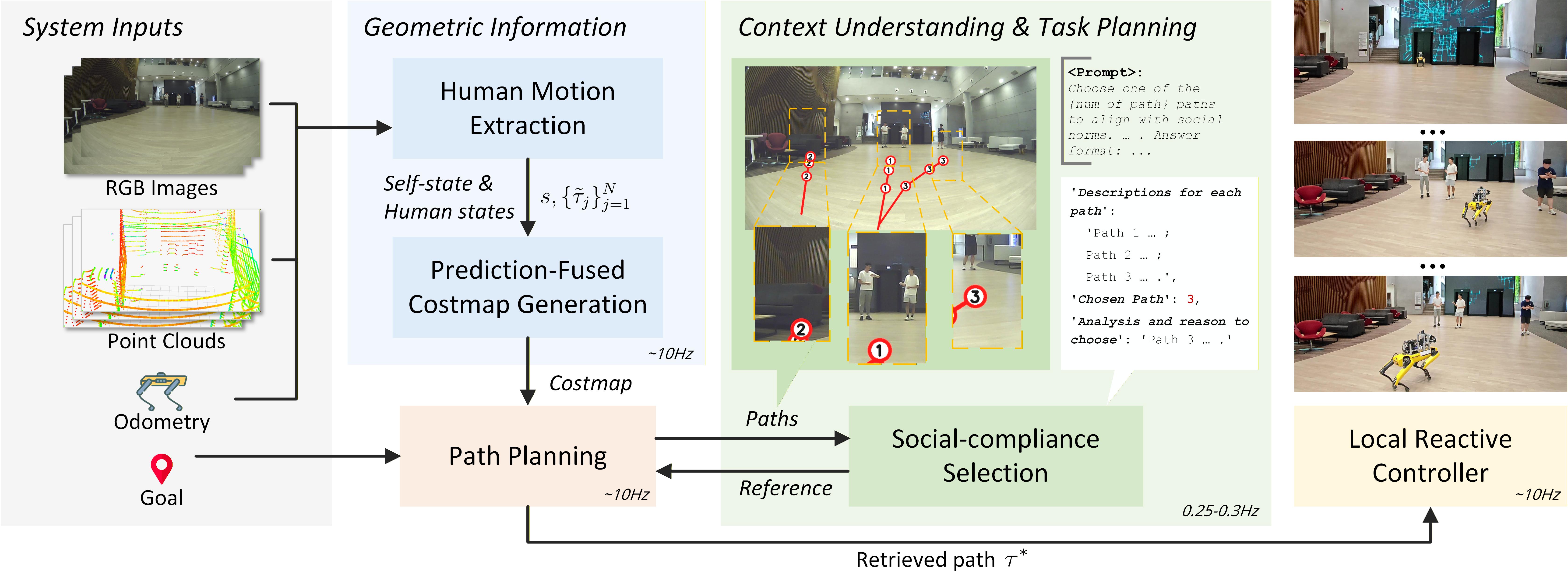}
  \vspace{-1em}
  \caption{System overview. Geometric constraints are extracted from human motion and costmap modules using sensor data. Collision-free path candidates are sampled, projected into the image, and evaluated by a fine-tuned VLM. The selection is fed back as reference to retrieve a path for the local controller.}
  \vspace{-1.5em}
  \label{fig:system_overview}
\end{figure*}


\section{Related Work}

\subsection{Navigation in Crowds}
Collision-free robot navigation has been extensively studied, from classical methods to reactive approaches such as ORCA for multi-agent interactions~\cite{van2011reciprocal}. Cooperative planners extend reciprocal assumptions to dense crowds, e.g., joint human-robot trajectory prediction with interacting Gaussian processes~\cite{trautman2015robot} and mixed-strategy Nash equilibrium models~\cite{muchen2024mixed}. Model-based approaches integrate Model Predictive Control (MPC) with additional elements such as Control Barrier Functions~\cite{vulcano2022safe}, topological invariance in cost functions~\cite{mavrogiannis2022winding}, or bilevel optimization with closed-loop crowd prediction~\cite{kucner2017enabling, samavi2024sicnav}. Learning-based methods leverage deep reinforcement learning with free-space prediction~\cite{sathyamoorthy2020densecavoid}, human-robot interaction features via local maps~\cite{chen2019crowd}, or interaction modeling with RNNs~\cite{liu2021decentralized} and attention~\cite{liu2023intention}. These methods mainly address multi-agent collision avoidance, optimizing safety, clearance, and travel time with limited consideration of social context. This work focuses on enabling robots to adapt to social contexts by analyzing human activities and relationships to guide behavior.

\subsection{Social Robot Navigation without Foundation Models}
Although social robot navigation has been studied for years, the definition of `social' remains ambiguous until a recent survey clarified it by proposing eight evaluative principles~\cite{francis2025principles}. Despite this progress, translating these principles into concrete costs or metrics remains difficult. Recent works increasingly explore robot social navigation capabilities, employing approaches such as manually designed social costs~\cite{triebel2016spencer, chen2017socially, hirose2023sacson, hirose2024selfi, li2024multi}, integration of MPC into Reinforcement Learning~\cite{han2025dr}, and behavior cloning with extensive training data~\cite{fahad2020learning, raj2024rethinking}. Many of these methods evaluate social compliance primarily through safety and distance metrics, such as human proximity or alignment with human collision avoidance preferences via Likert-scale assessments. They focus largely on collision avoidance and lack a deeper understanding of full semantic context, limiting their ability to interpret implicit social rules in complex environments. Our work introduces a system that integrates a task-specific fine-tuned VLM, capable of understanding social context across diverse environments and human activities, going beyond simple collision avoidance.

\subsection{Utilizing Foundation Models for Robot Navigation}
Foundation models like Large Language Models (LLMs) and Vision-Language Models (VLMs) are increasingly integrated into robot navigation. For example, LLMs can facilitate interpreting open-vocabulary semantics and constructing scene graphs or maps~\cite{werby2024hierarchical, gu2024conceptgraphs} that benefit objective goal navigation. VLMs can encode the traversability of different terrains and environmental objects~\cite{sathyamoorthy2024convoi, zhang2024interactive} for motion planning and instruction following. The closest work to ours in the propose-and-select concept is VL-TGS~\cite{song2025vl}. In contrast to focusing on static scene semantics, our approach reasons over dynamic, implicit, agent-dependent contexts arising from human behaviors and human-robot interactions. It therefore applies different constraints and strategies and does not specify explicit behavior rules like~\cite{liang2025mosu}.

CoNVOI~\cite{sathyamoorthy2024convoi} queries a large VLM to pick labels aligned with a specific behavioral phrase, making it prompt-specific.
VLM-Social-Nav~\cite{song2024vlm} incorporates VLMs for social navigation by prompting to output abstract language descriptions, which are then grounded into actions with heuristics, i.e., cost functions in a DWA controller. It focuses on immediate actions and lacks planning over a longer horizon. GSON~\cite{luo2025gson} addresses the problem by querying for group clustering, but it overlooks social norms in situations where explicit groups are absent. OLiVia-Nav~\cite{narasimhan2024olivia} extracts social context using a CLIP-based encoder and aligns embeddings with expert trajectories; however, this coarse alignment does not generalize well to more complicated scenes. In contrast, our proposed navigation structure integrates contextual understanding from visual cues and task-level planning into a fine-tuned VLM module to achieve effective socially compliant navigation and also generalization across diverse and unseen scenarios.


\section{Overview}

\subsection{Formulation}
The social robot navigation task can be conceptually understood as a multi-objective optimization problem, inspired by~\cite{mavrogiannis2023core, francis2025principles}. Consider a robot $r$ and a planar workspace $\mathcal{W} \subseteq \mathbb{R}^2$ and the traversable subspace is defined as $\mathcal{W}_{obs}^C = \mathcal{W} \setminus \mathcal{W}_{obs}$, where $\mathcal{W}_{obs} \subseteq \mathcal{W}$ is the subspace occupied by static obstacles. Assume that robot $r$ now lies in a configuration $s \in \mathcal{Q} \subseteq \mathit{SE}(2)$, with a mapping $\mathit{M}$ embedding $s$ into the workspace $\mathit{M}(s) \subseteq \mathcal{W}$ such that $q\notin\mathcal{Q}_{obs}$ $\forall q\in\mathcal{Q}$ when $\mathit{M}(q)\notin\mathcal{W}_{obs}$. During the normalized time interval $[0,1]$, the total number of human agents is $N$, with the individual path of each agent denoted as $\{\tau_j\}_{j=1}^N$, and the estimation of the path represented as $\{\tilde{\tau}_j\}_{j=1}^N$ since it is challenging to acquire the ground truth of the human agent behaviors. During planning, the robot computes a path $\tau \in \mathcal{T}$ toward the goal $g$ formulated as the following optimization problem:
\begin{equation}
    \begin{aligned}
        \tau = \argmin_{\tau\in\mathcal{T}}~~&\bigl( c(\tau), c^{social}(\tau, \{\tilde{\tau}_j\}_{j=1}^N; \rho_{crowd}) \bigr) \\
        \text{s.t.}\quad & \tau(t)\in\mathcal{W}_{obs}^C, \ \forall t\in[0,1], \\
            & \tau(t)\cap \tilde{\tau}_j(t) = \emptyset, \ \forall j\in \{1,..., N\}\, \ \forall t\in[0,1], \\
            & \tau(0) = \mathit{M}(s), 
            \tau(1) = \mathit{M}(g)
    \end{aligned}
    \label{eq:formulation}
\end{equation}
where $c: \mathcal{\tau} \rightarrow \mathbb{R}$ is the cost function for path specifications (e.g., time to goal) and environment constraints (e.g., untraversable obstacles). $c^{social}$ is a joint cost that combines all social considerations, whose form may vary with the density of the crowd $\rho_{crowd}$. The path $\tilde{\tau}_j$ can be expressed as explicit waypoints in a planar space or implicit feature representations. In this work, we focus on scenarios that involve more analysis of social norms beyond the collision avoidance under low to moderate crowd density, and omit $\rho_{crowd}$ for simplicity in the following context. Since human agents’ path estimations should be updated based on information collected up to the current time, the path optimization is performed in a receding horizon manner starting from the current time step, $t'$. The planning horizon is defined as $H$ with the constraint expressed as $\tau(t)\cap \tilde{\tau}_j(t) = \emptyset, \ \forall j\in \{1,..., N\}, \ \forall t\in[t',t'+H]$.
We assume the objective is decomposable into geometric feasibility and in-context semantics, with semantic optima forming a well-covered subset of the geometric-optimal set.

\begin{figure*}[tb]
  \centering
  \includegraphics[width=0.92\linewidth]{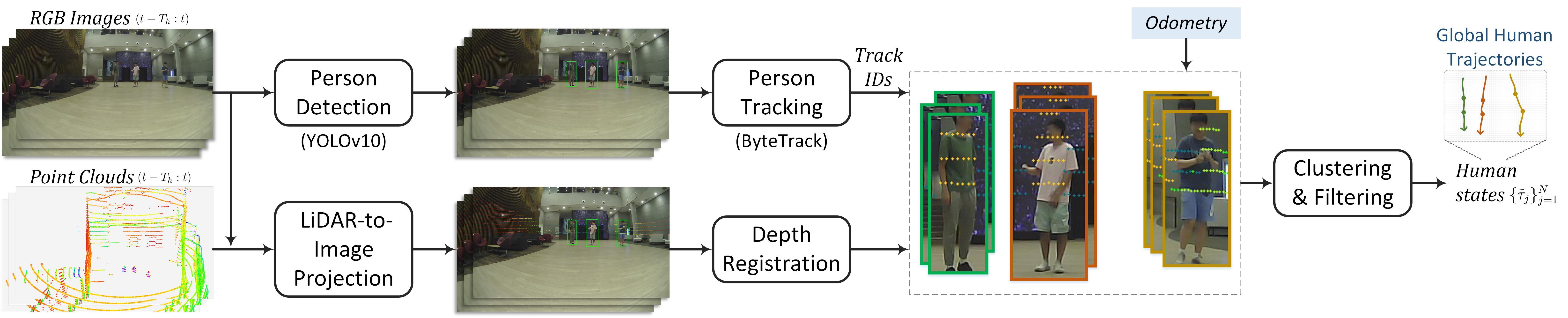}
  \vspace{-1em}
  \caption{Human Motion Extraction. The module detects and tracks humans using images, then fuses depth information from LiDAR point clouds with ego-pose from odometry to estimate human states in global coordinates.}
  \vspace{-1.5em}
  \label{fig:human_motion}
\end{figure*}

\subsection{System Design}
The system is built upon the formulated optimization problem. The core idea is to first generate plausible path candidates that inherently satisfy geometric constraints derived from sensor data. A VLM then selects the most socially compliant path. This enables simultaneous semantic scene understanding and task planning. Fig.~\ref{fig:system_overview} illustrates the overall navigation pipeline and the interactions between its modules.

Experimentally, we handle the constraints $\tau(t) \in \mathcal{W}_{obs}^C$ and $\tau(t) \cap \tilde{\tau}_j(t) = \emptyset$ using a costmap that fuses real-time obstacle sensing with short-horizon human motion predictions. Environmental constraints are assumed to be largely captured by the sensed costmap. The path specification part in the cost function $c(\tau)$ is absorbed directly into the sampling and post-processing of candidate paths. Human Motion Extraction (Sec.~\ref{sec:extract}) and Prediction-Fused Costmap Generation (Sec.~\ref{sec:costmap}) process incoming sensor data to obtain 3D human trajectories and generate future-aware costmaps with trajectory augmentation. Path Planning (Sec.~\ref{sec:pathplan}) then samples collision-free path candidates based on this costmap. Since the costmap encodes predicted human motions, distance-related factors in the social cost $c^{social}$ such as personal space and potential future blocking are partially addressed during path generation.

Subsequently, the Social-Compliance Selection module (Sec.~\ref{sec:select}) identifies the optimal path through a multiple-choice decision process. Acting as a proxy for the average human, the VLM assesses whether each candidate path adheres to social norms by jointly considering all factors in $c^{social}$, thereby incorporating social cost into its analysis and solve the minimization problem $\arg\min{ \bigl(c(\tau), c^{social}(\tau, \{\tilde{\tau}_j\}_{j=1}^N) \bigr)}$ by selecting the highest-scoring path. The Path Planning module replans regularly and retrieves the closest path upon receiving a reference path from the VLM output. Finally, the Local Reactive Controller (Sec.~\ref{sec:controller}) chooses suitable waypoints from the retrieved path as continuously updated sub-goals as the robot moves toward its final goal, and issues corresponding velocity commands while considering human states.

\section{Social Navigation with Path Selection}
\subsection{Human Motion Extraction}
\label{sec:extract}
Given a frame of LiDAR point clouds and an RGB image, we extract 3D human spatial information through detection, tracking, and depth registration. Instead of performing 3D object detection directly on point clouds where training data are largely drawn from autonomous driving and domain shifts can degrade performance, we apply the real-time 2D detector YOLOv10~\cite{wang2024yolov10} to RGB images to detect humans. Tracking is handled by ByteTrack~\cite{zhang2022bytetrack}, with tracklets buffered over the most recent 40 frames. All of these steps operate within the 2D image domain.

Depth registration is performed by fusing point clouds with image-based detections. As shown in Fig.~\ref{fig:human_motion}, LiDAR points are projected onto the image plane to generate a depth map. For each detected bounding box, we create a mask and apply KMeans clustering with three clusters to filter out background points and handle partial occlusions. The dominant cluster is then used to estimate the 3D position by averaging its points, which provides the human–robot distance. Human motion is categorized into two states: dynamic humans represented by sequences of 3D coordinates and velocities, and pseudo-static humans represented by a single 3D position for newly assigned track IDs without history. To reduce data association errors such as ID switches or mismatches that can distort frame-wise velocity estimation, we use a local temporal window with a maximum velocity constraint to smooth trajectories and remove outliers.

\subsection{Prediction-Fused Costmap Generation}
\label{sec:costmap}
\begin{figure}[tb] 
  \centering
  \includegraphics[width=0.9\linewidth]{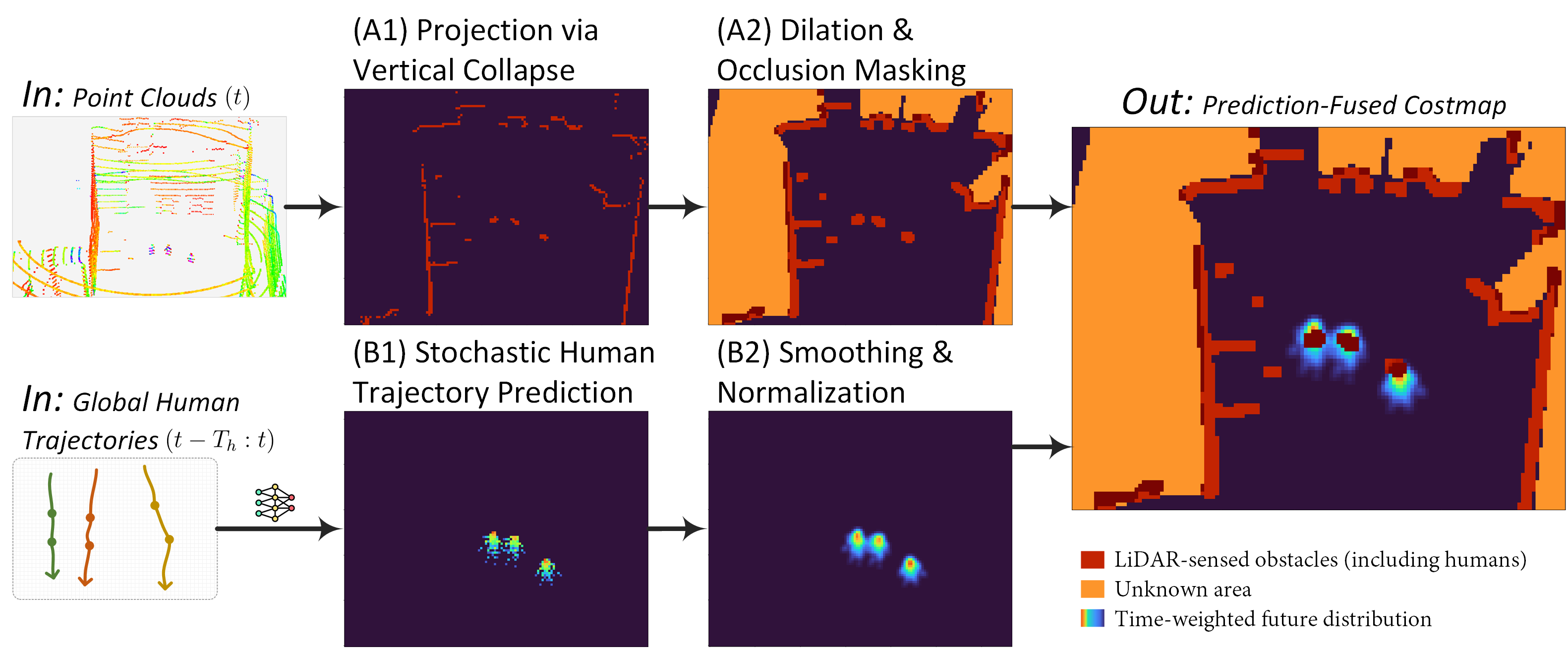}
  \vspace{-0.5em}
  \caption{The illustration of Prediction-Fused Costmap Generation.
  }
  \vspace{-1em}
  \label{fig:costmap}
\end{figure}

We construct a costmap that integrates real-time LiDAR-based obstacles with predicted human trajectories for path planning. This process projects the temporal dimension of human dynamics into the grid cost map. The module takes the dynamic human output from the Human Motion Extraction component as input. 
Fig.~\ref{fig:costmap} presents the iterative costmap generation process, which starts by projecting the 3D point cloud onto the ground plane. Any grid cell with vertical presence along the $z$-axis is marked as occupied. In parallel, a state-of-the-art stochastic human trajectory prediction network~\cite{fang2025neuralized} predicts human movements over a 2.4-second horizon and outputs 20 trajectory samples. Since the network models interaction dynamics among agents through a neuralized Markov Random Field, we add the ego-robot as a graph node to capture potential cooperative behaviors. Only human predictions are used for costmap generation. The predicted positions receive temporal decay weights from $0.8$ to $0.1$ and are aggregated by bin counting to assign grid costs. A Gaussian filter is then applied for smoothing. The prediction-based costmap is overlaid with the real-time LiDAR obstacle map to generate the final cost map.

\subsection{Path Planning}
\label{sec:pathplan}
\begin{figure}[tb]  
  \centering
  \includegraphics[width=0.95\linewidth]{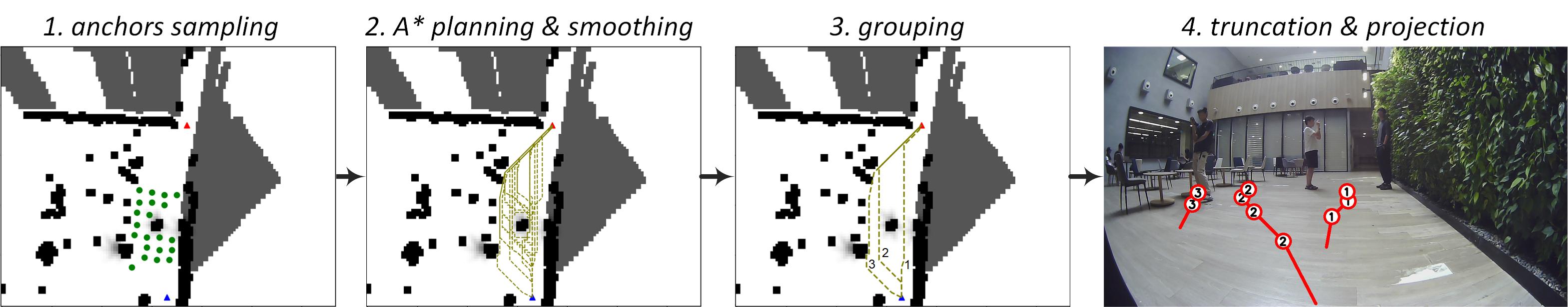}
  \vspace{-0.5em}
  \caption{Path Planning. Detoured yet collision-free path candidates are mainly obtained through the use of anchors.}
  \vspace{-1.5em}
  \label{fig:path_plan}
\end{figure}

The path planning module is based on the classical A* algorithm to generate multiple feasible paths from the position of the robot to a specified goal. As depicted in Fig.~\ref{fig:path_plan},  we introduce strategies to diversify candidates since standard A* returns only the shortest path. In practice, socially compliant paths are often longer as they may involve yielding or detours. To model such behaviors, we introduce anchors that guide paths through intermediate points to promote variations that reflect potential human-robot interactions. Anchors are generated within a rectangular region directed from the ego position toward the goal with a predefined width. We adopt the Poisson disk sampling principle to ensure well-spaced anchors, which random sampling cannot guarantee. To simplify implementation, the rectangular region is discretized into a grid with a defined cell size, and grid corners sufficiently distant from obstacles are selected as anchors based on the cost map. Each path is then generated by applying A* from the start to an anchor and then from the anchor to the goal, concatenating both segments. Multiple anchors produce diverse path modalities aligned with social navigation behaviors. To reduce redundancy, we cluster paths using a Hausdorff distance~\cite{raj2024rethinking} threshold and retain three to six representatives. These candidates are then projected onto the RGB image for input to the VLM. This module continuously replans candidate paths as the robot moves. Due to VLM inference latency, the selected path serves only as a reference and is not directly followed. Upon receiving the VLM output, the system retrieves the closest matching path from the current candidate set, while the local reactive controller handles velocity planning and dynamic obstacle avoidance.

\subsection{Social-compliance Selection}
\label{sec:select}
Our work grounds social compliance in selecting a path that minimizes interruptions to ongoing human activities. We leverage VLMs for semantic understanding to provide high-level path reference, inferring likely human motions and interactions from posture and context to reason about social norms along each path, while decoupling path selection from velocity control for asynchronous operation.

\subsubsection{Visual Prompting}
Given an input image with multiple projected paths and depth information, a large VLM such as GPT-4o can interpret the scene, evaluate each path according to social norms, and select the most appropriate option. This evaluation can also be performed by a smaller fine-tuned VLM such as Qwen-2.5 at inference, using a similar constrained prompting strategy.
To encourage socially compliant reasoning, prompts include (i) a one-sentence description of each candidate path, (ii) an explicit choice of the selected path, and (iii) a justification of the decision, all within a structured output format as shown in Fig.~\ref{fig:vlm}. By constraining the output format, we encourage the model to explicitly generate chain-of-thought reasoning, thereby enhancing complex reasoning performance. GPT-4o receives richer instructions to elicit social-norm analysis, while Qwen-2.5 uses simplified prompts and RGB input only to reduce latency while preserving the output schema.

\subsubsection{Fine-tuning}
\begin{figure}[!tb]
  \centering
  \includegraphics[width=0.98\linewidth]{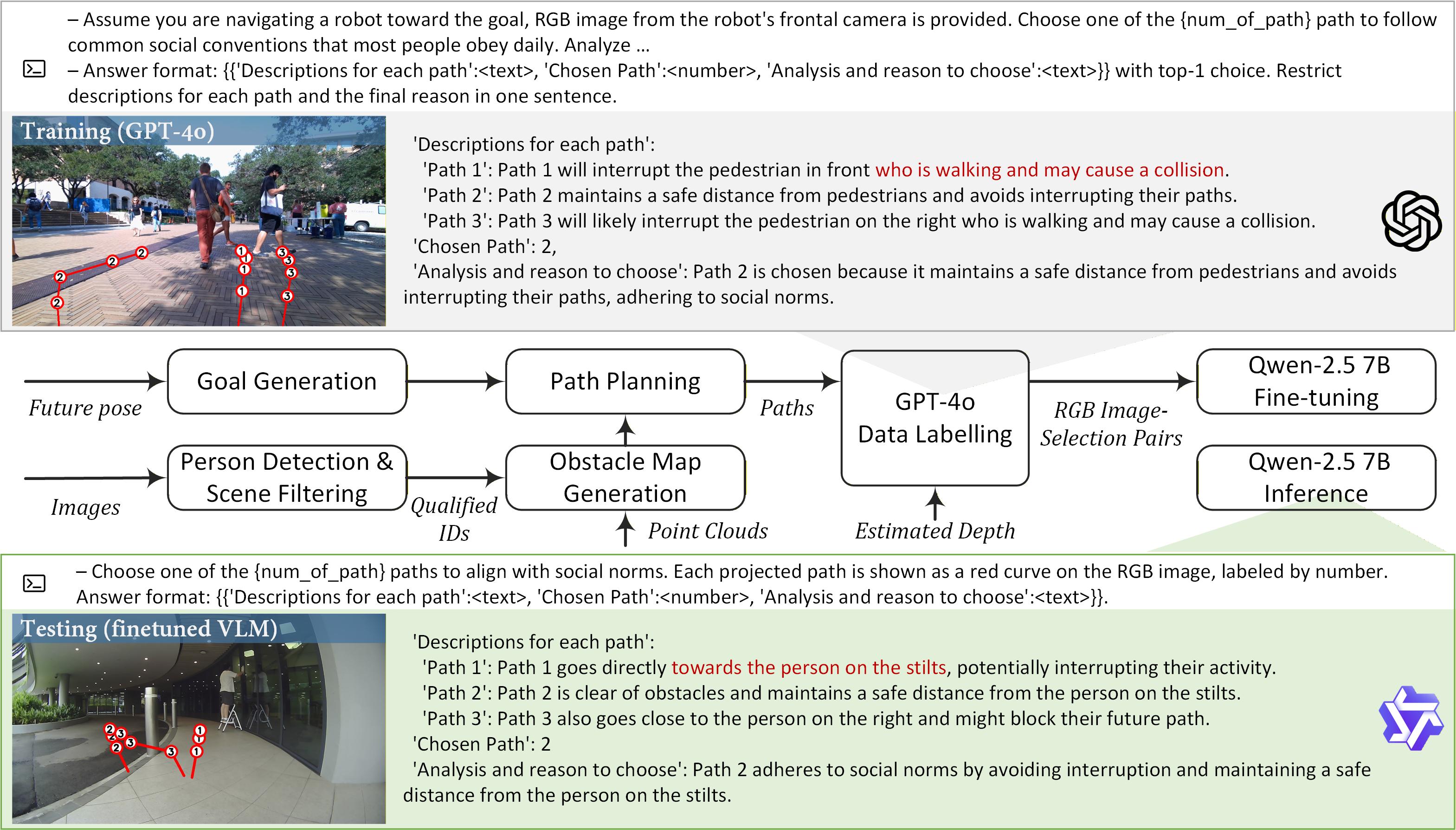}
  \vspace{-1em}
  \caption{Fine-tuning data generation and example query-answer pairs.}
  \vspace{-1.5em}
  \label{fig:vlm}
\end{figure}

While GPT-4o demonstrates strong performance, its high inference latency makes it impractical for real-time decision-making. To address this limitation, our objective is to distill GPT-4o’s socially compliant reasoning ability into a smaller and faster VLM, Qwen-2.5 7B~\cite{bai2025qwen2}, which offers significantly lower latency. This enables robots to respond more rapidly to dynamic environmental changes while maintaining social awareness. 
Specifically, we use the SCAND~\cite{karnan2022socially} dataset to generate training data for fine-tuning the Qwen model, as illustrated in Fig.~\ref{fig:vlm}. For each frame, a rough goal direction is estimated from the future $6$-second trajectory and then perturbed within $\pm 10^\circ$ to sample a goal location $13$ to $14.5$ meters away. Due to the absence of LiDAR–camera extrinsics, we run our path planning algorithm on instantaneous obstacle maps from Velodyne Puck point clouds and project the resulting paths onto the image using approximate calibration. The dataset is downsampled to $1$ Hz and frames without humans are filtered out, yielding $4{,}851$ image–selection pairs for training and $574$ for validation.

As Qwen-2.5 7B model already possesses strong general visual reasoning abilities, our fine-tuning aims to constrain its output format and stimulate its inherent social reasoning capabilities beyond the dataset. We perform full-model supervised fine-tuning using two H100 GPUs. At inference, the model is served with vLLM~\cite{kwon2023efficient} in 4-bit quantization for efficiency: input images are sent to the H100 server, which processes them and returns the output to the robot. 

\subsection{Local Reactive Controller}
\label{sec:controller}
The velocity control module in our pipeline builds on the ORCA algorithm: the set of permitted velocities $ORCA^\tau_{A|B}$ for agent $A$ to avoid collision with agent $B$ within time $\tau$ is $\{\mathbf{v} |(\mathbf{v}- (\mathbf{v}^{opt}_A +\frac{1}{2}\mathbf{u})) \cdot \mathbf{n}\geq 0\}$, where $\mathbf{v}^{opt}_A$ is the adopted velocity of $A$ and $\mathbf{u}$ is the smallest required change. The factor $\tfrac{1}{2}$ denotes the equal responsibility shared between agents $A$ and $B$. We adapt this shared-responsibility concept by introducing a velocity vector–guided scheme, which allows the robot to assume greater responsibility when the approach direction of the pedestrian aligns with the line connecting the two agents. The robot’s responsibility is assigned in a piecewise manner: it increases linearly with $\theta$ when the angle $\theta$ between the position and velocity differences is within $(0, \frac{\pi}{2})$, and is fixed at one-half otherwise, while retrieved path waypoints serve as sub-goals for velocity generation.

\begin{figure*}[tb]
  \centering
  \includegraphics[width=0.93\linewidth]{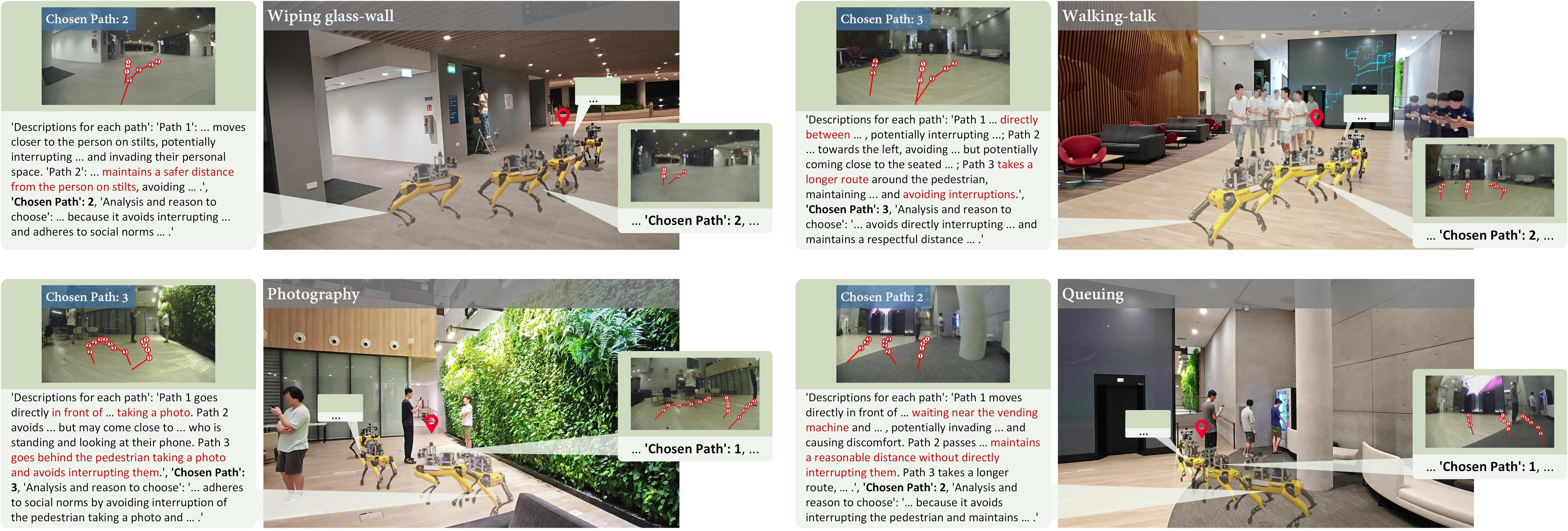}
  \vspace{-1em}
  \caption{Navigation performance on four experimental scenarios. Without any consideration of collision avoidance and social compliance, the goal is set directly ahead of the starting point, marked by a red map pin. For each scenario, we display and highlight one VLM-generated answer to the image query on the left-hand side.}
  \vspace{-1em}
  \label{fig:exp_results}
\end{figure*}

\begin{table*}[tb]
    \centering
    \caption{Navigation performance comparison on four different scenarios.}
    \label{tab:exp_result}
    \vspace{-1em}
    \begin{minipage}{0.49\linewidth}
        \centering
        \textbf{(a) Wiping glass-wall}\vspace{0.2em} \\
        \setlength{\tabcolsep}{4pt}
        \begin{tabular}{lccccc}
        \toprule
            Method         & NT$\downarrow$ & PSV$\downarrow$ & TFP$\downarrow$ & SIT$\downarrow$ & Max. SIR$\downarrow$ (\%) \\
        \midrule
            G-MPC          & 54.55 & 0.00 & 0.00 & 8.41 & 13.22 \\
            AttnGraph-RL   & 29.11 & 3.81 & 1.75 & 5.65 & 16.97 \\
            ViNT           & 46.24 & 0.00 & 0.00 & 12.32 & 16.97 \\
            VLM-Social-Nav & 32.21 & 0.00 & 0.00 & 0.71 & 2.86 \\
            GSON           & 30.42 & 0.00 & 0.00 & 8.28 & 16.97 \\
            \rowcolor{gray!20}
            \textbf{Ours}  & \textbf{28.48} & \textbf{0.00} & \textbf{0.00} & \textbf{0.00} & \textbf{0.00} \\
        \bottomrule
        \end{tabular}
    \end{minipage}
    \hfill
    \begin{minipage}{0.49\linewidth}
        \centering
        \textbf{(b) Walking-talk}\vspace{0.2em} \\
        \setlength{\tabcolsep}{4pt}
        \begin{tabular}{lccccc}
        \toprule
            Method         & NT$\downarrow$ & PSV$\downarrow$ & TFP$\downarrow$ & SIT$\downarrow$ & Max. SIR$\downarrow$ (\%) \\
        \midrule
            G-MPC          & 68.13 & 1.73 & 5.69 & 2.63 & 44.10 \\
            AttnGraph-RL   & *90.00 & 0.20 & \textbf{1.66} & 0.00 & 0.00 \\
            ViNT           & 47.14 & 1.54 & 4.01 & 1.43 & 42.80 \\
            VLM-Social-Nav & *90.00 & 4.99 & 7.57 & 0.00 & 0.00 \\
            GSON           & 38.25 & 0.00 & 6.59 & 1.35 & 39.70 \\
            \rowcolor{gray!20} 
            \textbf{Ours}  & \textbf{28.49} & \textbf{0.00} & 1.80 & \textbf{0.00} & \textbf{0.00} \\
        \bottomrule
        \end{tabular}
    \end{minipage}
    
    \vspace{0.5em}
    
    \begin{minipage}{0.49\linewidth}
        \centering
        \textbf{(c) Photography}\vspace{0.2em} \\
        \setlength{\tabcolsep}{4pt}
        \begin{tabular}{lccccc}
        \toprule
            Method         & NT$\downarrow$ & PSV$\downarrow$ & TFP$\downarrow$ & SIT$\downarrow$ & Max. SIR$\downarrow$ (\%) \\
        \midrule
            G-MPC          & 36.14 & 3.68 & 8.07 & 3.67 & 27.66 \\
            AttnGraph-RL   & *90.00 & 1.71 & 5.28 & 0.00 & 0.00 \\
            ViNT           & 36.82 & 2.91 & 9.28 & 3.43 & 27.66 \\
            VLM-Social-Nav & \textbf{27.01} & 7.15 & 13.03 & 3.89 & 26.46 \\
            GSON           & 46.54 & 2.41 & 5.72 & 0.68 & 6.77 \\
            \rowcolor{gray!20} 
            \textbf{Ours}  & 40.15 & \textbf{0.0} & \textbf{4.32} & \textbf{0.00} & \textbf{0.00} \\
        \bottomrule
        \end{tabular}
    \end{minipage}
    \hfill
    \begin{minipage}{0.49\linewidth}
        \centering
        \textbf{(d) Queuing}\vspace{0.2em} \\
        \setlength{\tabcolsep}{4pt}
        \begin{tabular}{lccccc}
        \toprule
            Method         & NT$\downarrow$ & PSV$\downarrow$ & TFP$\downarrow$ & SIT$\downarrow$ & Max. SIR$\downarrow$ (\%) \\
        \midrule
            G-MPC          & 47.11 & 5.22 & 21.26 & 20.26 & 20.46 \\
            AttnGraph-RL   & \textbf{17.18} & 1.25 & 5.45 & 2.81 & 21.03 \\
            ViNT           & *90.00 & 1.45 & 6.25 & 0.00 & 0.00 \\
            VLM-Social-Nav & *90.00 & \textbf{0.00} & 4.25 & 0.00 & 0.00 \\
            GSON           & 29.10 & 0.38 & \textbf{2.36} & 0.00 & 0.00 \\
            \rowcolor{gray!20} 
            \textbf{Ours}  & 26.14 & 0.33 & 2.70 & \textbf{0.00} & \textbf{0.00} \\
        \bottomrule
        \end{tabular}
    \end{minipage}
    
\vspace{0.3em}
\begin{minipage}[t]{0.96\textwidth}
    \footnotesize Numbers reported are the average of three trials. \\
    \footnotesize *If the robot fails to reach the goal within 90 seconds, we note the navigation time (NT) as 90. 
\end{minipage}
\vspace{-2em}
\end{table*}

\section{Experiments}
\subsection{Platform Description}
Our implementation is based on a Boston Dynamics Spot legged robot, equipped with an NVIDIA Jetson Orin for onboard navigation processing, excluding the VLM. The fine-tuned Qwen model runs as a network service on an H100 GPU. The sensing inputs include monocular RGB images from an ELP fisheye camera and point clouds from a Hesai JT16 mechanical LiDAR. Odometry data is obtained directly from the robot.

\subsection{Experimental Setup}
We evaluate our proposed navigation pipeline with five other baselines via four controlled scenarios to address social contexts. Agents are asked to stand at pre-designed positions or walk along a pre-designed direction to imitate multiple social situations encountered in daily life.

\textit{Experimental scenarios}.
Drawing on insights from social scenarios and previous works~\cite{song2024vlm,luo2025gson,francis2025principles}, we design four experimental settings with an expected robot behavior, respectively:
(a) \textbf{Wiping glass-wall} -- A single person wipes a glass wall while standing on a ladder; the robot is expected to take a wide detour and maintain a safe distance to account for potential fall hazards.
(b) \textbf{Walking-talk} -- Two people walk while conversing and accompanied by another person walking alone; the robot is expected to avoid passing through the gap between the conversing individuals.
(c) \textbf{Photography} -- Two individuals take photographs while a third person stands nearby using a mobile phone; the robot is expected to avoid crossing the line of sight between the photographer and the model.
(d) \textbf{Queuing} -- Multiple individuals queue at a vending machine; the robot is expected to not cut in line.

\textit{Baselines}.
Social compliance is compared with five different navigation benchmarks, with 3 trials per scenario:
\begin{itemize}
    \item \textbf{G-MPC}~\cite{wang2022group} which leverages group-based prediction for the MPC controller and achieves fewer social group intrusions.
    \item \textbf{Attention-Based Interaction Graph}~\cite{liu2023intention} (referred to as AttnGraph-RL in the following sections), an RL framework demonstrates the robot’s social awareness for not intruding into the intended paths of other agents.
    \item \textbf{ViNT}~\cite{shah2023vint}, a foundation model of visual navigation pretrained on multiple large-scale navigation datasets, including SCAND for social navigation.
    \item \textbf{VLM-Social-Nav}~\cite{song2024vlm} that integrates the VLM for abstract motion scoring.
    \item \textbf{GSON}~\cite{luo2025gson} which utilizes the VLM for social group identification and navigates to avoid.
\end{itemize}
We follow the official implementations if the code is publicly available. For VLM-Social-Nav, we deploy the planning algorithm based on the original paper and its default parameter setting, while the perception module is realized using our method described in Sec.~\ref{sec:extract}.

\textit{Metrics}.
We use five metrics to evaluate robot behavior, three of which follow prior studies: 
(i) Navigation Time (\textbf{NT}) -- Total time to reach the goal. 
(ii) Personal Space Violation duration (\textbf{PSV}) -- Time spent within $0.25$ m of any person~\cite{narasimhan2024olivia, hirose2024selfi}. 
(iii) Time Facing Pedestrians (\textbf{TFP}) -- Duration of the robot facing a pedestrian~\cite{kastner2024arena}. We adopt a conservative definition, counting only within a $30^\circ$ cone and a range of $3$m. Two new metrics are also introduced to assess social-zone disruptions: 
(iv) Social-zone Interruption Time (\textbf{SIT}) -- Total time the robot occupies any defined social zone. 
(v) Maximum Social-zone Interruption Ratio (\textbf{Max. SIR}) -- The largest ratio of the overlapped area of Spot’s footprint ($0.5 \times 1.1$ m) to the area of any social zone. The social zone in scene (a) is defined as a square with a side length of $1.8$m based on fall-safety guidelines. In other scenes, it is defined as the area between people engaged in conversation, photography, or queuing.

\subsection{Experiment Results}
Fig.~\ref{fig:exp_results} visualizes the navigation process of our system, and Table~\ref{tab:exp_result} presents the quantitative comparisons of the five evaluation metrics. The full navigation behaviors for all approaches are available in the supplementary video. Our method delivers consistently strong performance across all scenarios. It achieves collision-free navigation without intruding into social zones and maintains low duration of personal space violation (PSV) and time facing pedestrians (TFP). This performance is attributed to our two-stage strategy: sampling paths that respect both human motion dynamics and obstacle constraints, followed by social context reasoning and task planning through a fine-tuned VLM with a high-frequency low-level controller using the planned path as a reference. The design enables dynamic adaptation to environmental changes, ensures compliance with social norms, and generalizes effectively across diverse contexts. Additionally, predictive future path referencing helps mitigate VLM-induced latency.
In contrast, VLM-Social-Nav scores actions solely on step-wise optimality. This approach can produce suboptimal behaviors in multi-human scenarios. Although its performance matches ours in the \textit{Wiping glass-wall} scenario, it shows undesirable outcomes in others. In the \textit{Walking-talk} scenario, it blocks a pedestrian by stopping and rotating in front of them. In the \textit{Queuing} scenario, it turns right repeatedly before becoming stuck in front of an obstacle. These behaviors may lower social-zone intrusion but prevent the robot from reaching the goal within the time limits as shown in Table~\ref{tab:exp_result} (b) and (d).

Another baseline GSON performs well in most static and group-based scenarios but struggles in dynamic interactions such as \textit{Walking-talk}, where it reacts too late after the talking agents have moved forward. In the static \textit{Photography} scene, as shown in Table~\ref{tab:exp_result} (c), it fails once and passes behind the posing person in two trials, which can still cause disturbance. It also fails in non–group contexts, such as the \textit{Wiping glass-wall} scenario where fall-down safety is critical. G-MPC fails in all four scenarios due to incorrect group identification, particularly when socially connected individuals are not spatially clustered. 
For other navigation approaches, ViNT often follows a straight-line path toward the goal, which frequently leads to social-zone intrusions despite being trained on the SCAND dataset. In the \textit{Queuing} scenario, it attempts a left turn to bypass the line but becomes stuck against a wall. AttnGraph-RL is the only method that shows unsafe behavior by colliding with the ladder in the \textit{Wiping glass-wall} scenario. Although it avoids talking and photographing groups as given in Table~\ref{tab:exp_result} (b) and (c), the method later deviates sharply and fails to reach the goal.

\begin{figure}[!tb]
  \centering
  \includegraphics[width=0.85\linewidth]{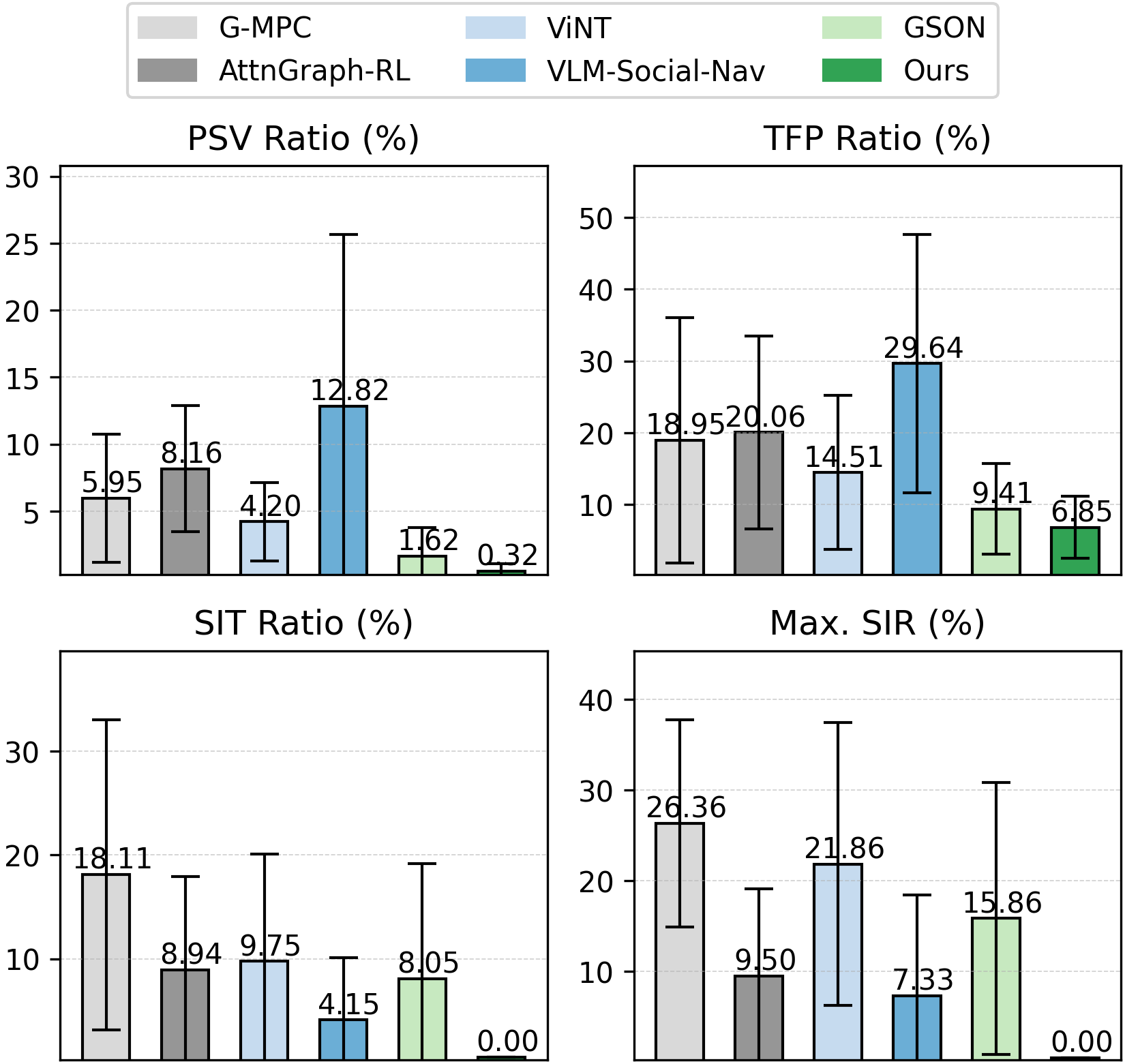}
  \vspace{-0.5em}
  \caption{Metrics comparison across four experimental scenarios. We use the time stamp of the human interruption as the total navigation time when the robot cannot reach the goal.}
  \vspace{-2em}
  \label{fig:social_metrics}
\end{figure}

In Fig.~\ref{fig:social_metrics}, we further compare the social compliance of each method across four scenarios. The sub-figures illustrate the proportion of time each method exhibits socially non-compliant behavior relative to the total navigation time. Our proposed method consistently achieves the lowest ratios across all metrics and scenarios, indicating robust adherence to social norms while still efficiently reaching the goal.

\subsection{Ablations}
We conduct ablations on two aspects: (i) the benefit of encoding temporal information versus instant action scoring as in VLM-Social-Nav, and (ii) the effect of motion prediction horizon. For (i), we use only the shortest visible path segment ($1.2$ m) for visualization and selection, approximating single-step direction selection. Paths are classified as straight if their angle with the vertical is $\leq 30^\circ$, and as left/right otherwise. Comparisons are based on snapshots taken when VLM-Social-Nav made its action decisions across all trials. As shown in Table~\ref{tab:ablation} (a), the selections closely match those of VLM-Social-Nav. This indicates the effectiveness of our planning approach, as these behaviors are validated in real-world experiments. For horizon analysis, we test $10$ scenes with ground-truth positions and group annotations. For each, we generate the prediction-fused costmap, specify start and goal, sample $36$ \textit{pre-clustering} paths with our planner, and count paths intersecting group regions as interruptions. Each horizon is evaluated over $50$ trials ($5$ runs per group), with average interruption ratios reported in Table~\ref{tab:ablation} (b). Longer prediction horizons reduce the generation of socially inappropriate paths, which in turn decreases the number surviving post-processing and results in a cleaner candidate set for the VLM.

\begin{table}
\centering
\caption{Ablation on path length and prediction horizon}
\label{tab:ablation}
\vspace{-1em}
    \begin{minipage}{\linewidth}
    \centering
    \textbf{(a) Shortest visible path}\\
    \setlength{\tabcolsep}{12pt}
    \begin{tabular}{lcccc}
    \toprule
        Scenarios    & (a) & (b) & (c) & (d) \\
    \midrule
        Path/Direction align. & 12/12 & 8/9 & 7/9 & 6/7 \\
    \bottomrule
    \end{tabular}
    \end{minipage}
    
    \vspace{0.5em}
    \begin{minipage}{\linewidth}
    \centering
    \textbf{(b) Prediction horizon}\\
    \setlength{\tabcolsep}{3pt}
    \begin{tabular}{lccccccc}
    \toprule
        Time(s)  & 2.4  &  2.0  &  1.6  &  1.2  &  0.8  &  0.4  & 0 \\
    \midrule
        Group Interrupt.(\%) & 8.89 & 12.50 & 15.83 & 21.11 & 26.39 & 29.44 & 39.17 \\
    \bottomrule
    \end{tabular}
    \end{minipage}
\vspace{-2em}
\end{table}

\section{Conclusion and Limitation}
We present a navigation pipeline that enables collision-free, socially compliant, and efficient robot navigation across diverse human-centered contexts. Our method first extracts human motion information from images and point clouds, then fuses motion predictions to guide collision-free path sampling under geometric constraints. A context-aware task planning module, powered by a fine-tuned VLM, interprets social norms and selects the best path accordingly. Extensive real-world experiments demonstrate that our system adapts robustly to various social scenarios, outperforming baseline methods across multiple social metrics. Additional ablation studies further validate the effectiveness of our design.

\textbf{Failure analysis}. Despite these results, potential failures remain, which can be grouped into three types: (i) no memory: the VLM sees only the current frame, missing social context that extends beyond the camera view (e.g., long queues); (ii) confounding activities: multiple simultaneous human actions or ambiguous intents can confuse the VLM’s prioritization of social norms; (iii) non-optimal candidates: post-processing may exclude the ideal path, leading either to minimal clearance or overly conservative detours.

\section*{APPENDIX}

\subsection{Additional Details}
As partially omitted in Fig.~\ref{fig:vlm}, the full prompt used for ground-truth labeling with GPT-4o in the fine-tuning process is provided below, while the prompt used during inference is fully shown in the figure:
\begin{mdframed}[
  backgroundcolor=gray!10,
  linecolor=gray!40,
  linewidth=0.5pt
]
\ttfamily\small
Assume you are navigating a robot toward the goal. An RGB image from the robot's
frontal camera is provided. Choose one of the \{num\_of\_path\} paths to follow
common social conventions that most people obey daily. \\
- Analyze each path along the red line and its labelled number, considering the current social context and human behaviors. You also have depth information from 0 to 255, where low depth means it’s further. \\
- Various norms may apply, but always prioritize the most essential ones. \\
- Double-check whether the selected path will interrupt pedestrians' activities near the path or enter an improper area. \\

Answer format: {{'Descriptions for each path':<text>, 'Chosen Path':<number>, 'Analysis and reason to choose':<text>}} with top-1 choice. Restrict descriptions for each path and the final reason in one sentence.
\end{mdframed}

\begin{figure}[H]
  \centering
  \includegraphics[width=0.92\linewidth]{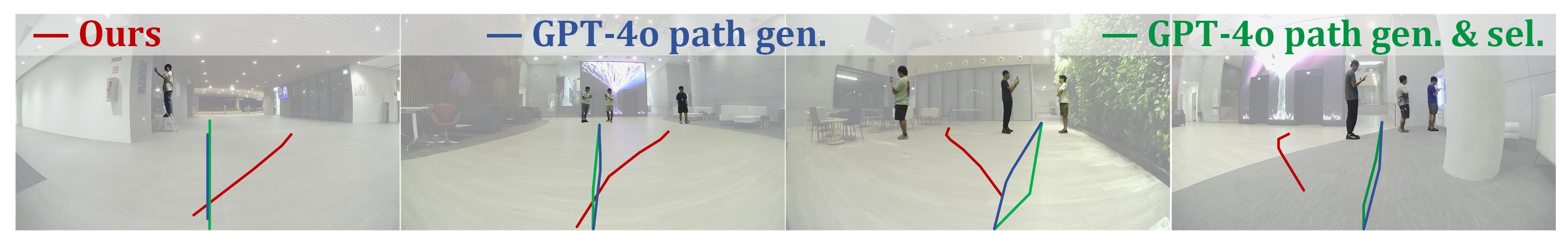}
  \vspace{-0.7em}
  \caption{Visualization of different path sampling strategies.}
  \label{fig:ablation}
\end{figure}

Beyond the ablations presented in the main text, we also evaluate the effect of decoupling geometrically constrained path sampling from contextual understanding and task planning. Specifically, we test the ability of the VLM to either directly generate a path or select the optimal one from multiple generated candidates. As depicted in Fig.~\ref{fig:ablation}, the VLM-generated paths deviate significantly from those produced by our method. They are often straight lines toward the goal without detours or any consideration of collision avoidance or social constraints.

\subsection{Human Study}
\begin{figure}[H]
  \centering
  \includegraphics[width=0.92\linewidth]{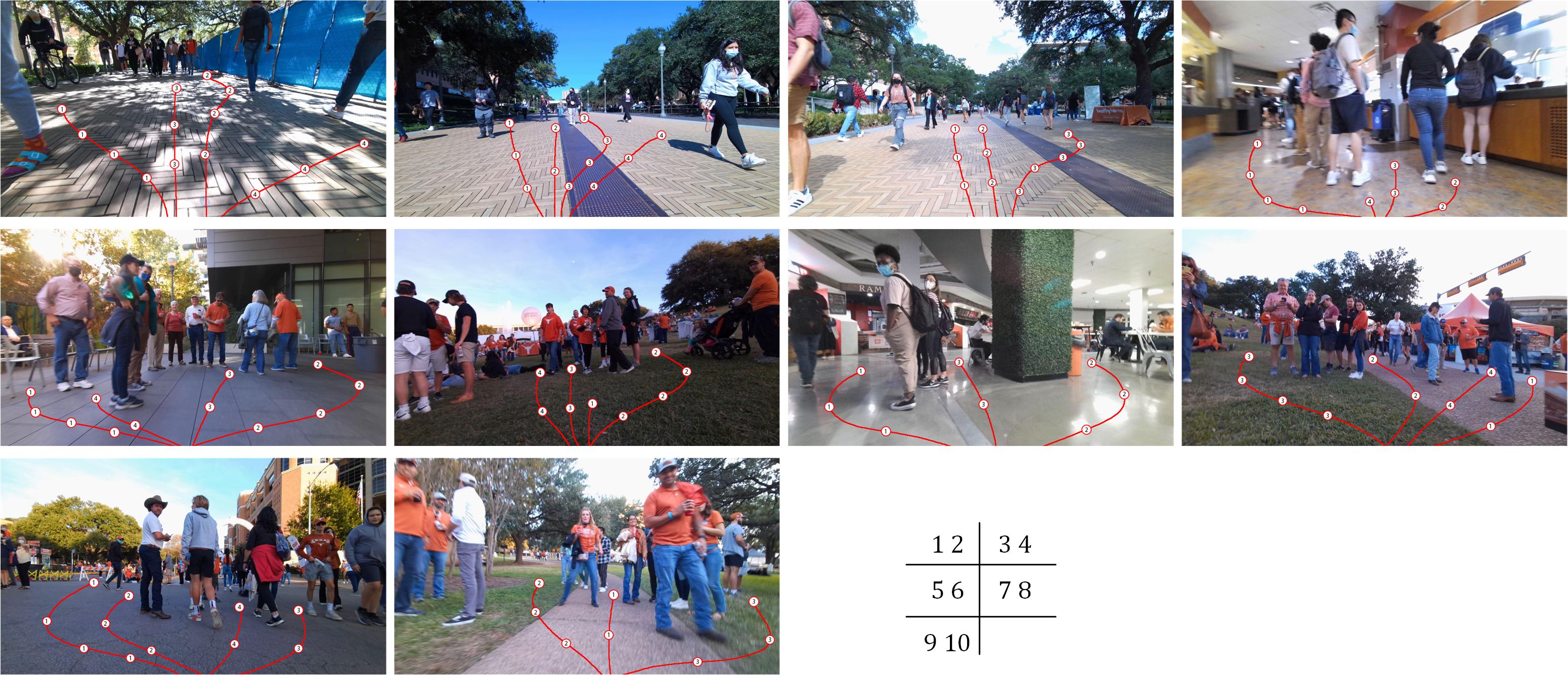}
  \vspace{-0.7em}
  \caption{Expert-drawn paths on sampled scenarios from the SCAND dataset~\cite{karnan2022socially}.}
  \label{fig:human_drawn}
\end{figure}

Before designing the complete system, we conducted a user study with $54$ participants to evaluate whether a VLM can reliably select socially compliant paths in various scenarios. Each image is annotated with $3-4$ feasible paths by human experts (see Fig.~\ref{fig:human_drawn}). Participants were asked to select one path aligned with their own daily walking preference while respecting social norms. We compare these selections with 12 VLM queries per image.

\begin{figure}[H]
  \centering
  \includegraphics[width=0.92\linewidth]{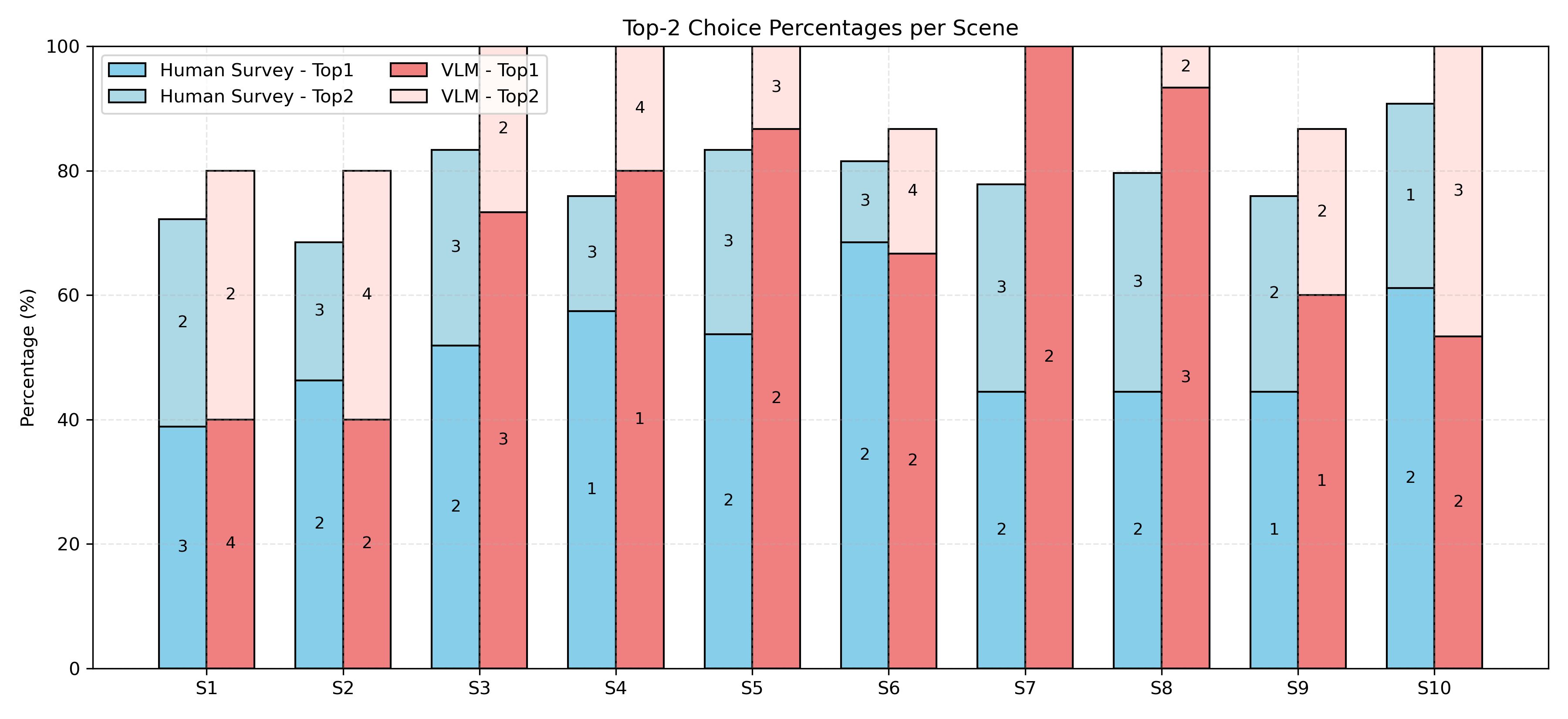}
  \vspace{-0.7em}
  \caption{Comparison between human and VLM path selections.}
  \label{fig:choice}
\end{figure}

Across the test scenarios, the VLM's Top-$2$ choices align with the dominant human modes in the majority of samples, as shown in Fig.~\ref{fig:choice}. Although human selections are inherently multimodal, the VLM consistently captures the primary or secondary socially normative patterns that appear across participants’ choices. This suggests that a VLM can generalize across diverse scenes and social contexts and serve as a semantic evaluator for distinguishing socially acceptable paths among geometrically feasible candidates.

\end{document}